\documentclass[conference]{IEEEtran}
\IEEEoverridecommandlockouts
\usepackage{cite}
\usepackage{amsmath,amssymb,amsfonts}
\usepackage{listings}
\usepackage{xcolor}
\usepackage{algorithmic}
\usepackage{graphicx}
\usepackage{textcomp}
\usepackage{xcolor}
\usepackage{booktabs}
\usepackage{tabularx}
\usepackage{array}
\usepackage{dcolumn}
\usepackage{units}
\usepackage{url}
\newcolumntype{Y}{>{\centering\arraybackslash}X}
\newcolumntype{R}{>{\raggedleft\arraybackslash}X}
\def\BibTeX{{\rm B\kern-.05em{\sc i\kern-.025em b}\kern-.08em
    T\kern-.1667em\lower.7ex\hbox{E}\kern-.125emX}}

\IEEEoverridecommandlockouts
\IEEEpubid{\makebox[\columnwidth]{\hfill © 2025 IEEE \hfill}%
\hspace{\columnsep}\makebox[\columnwidth]{ }}
    
\begin{document}
\title{An Agentic AI Pipeline for Appliance-Level Energy Anomaly Detection and LLM-Driven Recommendations}
\author{
\IEEEauthorblockN{Dihia Falouz}
\IEEEauthorblockA{\parbox[t]{2.15in}{\centering
\textit{Laboratoire LITAN} \\
\textit{École supérieure en Sciences et Technologies de l'Informatique et du Numérique}\\
RN 75, Amizour 06300, Bejaia, Algérie \\
d\_falouz@estin.dz}}
\and
\IEEEauthorblockN{Aida Douaibia}
\IEEEauthorblockA{\parbox[t]{2.15in}{\centering
\textit{Laboratoire LITAN} \\
\textit{École supérieure en Sciences et Technologies de l'Informatique et du Numérique}\\
RN 75, Amizour 06300, Bejaia, Algérie \\
a\_douaibia@estin.dz}}
\and
\IEEEauthorblockN{Amine Bechar}
\IEEEauthorblockA{\parbox[t]{2.15in}{\centering
\textit{Laboratoire LITAN} \\
\textit{École supérieure en Sciences et Technologies de l'Informatique et du Numérique}\\
RN 75, Amizour 06300, Bejaia, Algérie \\
bechar@estin.dz}}
\linebreak\and
\IEEEauthorblockN{Youssef Elmir}
\IEEEauthorblockA{\parbox[t]{2.15in}{\centering
\textit{Laboratoire LITAN} \\
\textit{École supérieure en Sciences et Technologies de l'Informatique et du Numérique}\\
RN 75, Amizour 06300, Bejaia, Algérie \\
elmir@estin.dz}}
\and
\IEEEauthorblockN{Abbes Amira}ء
\IEEEauthorblockA{\parbox[t]{2.15in}{\centering
\textit{College of Computing and Informatics} \\
\textit{University of Sharjah}\\
Sharjah, UAE \\
aamira@sharjah.ac.ae}}
\and
\IEEEauthorblockN{Adel Oulefki}
\IEEEauthorblockA{\parbox[t]{2.15in}{\centering
\textit{College of Engineering and Information Technology (CEIT)} \\
\textit{University of Dubai}\\
Dubai, UAE \\
aoulefki@ud.ac.ae}}
}
\maketitle
\begin{abstract}
Appliance-level energy monitoring in office buildings produces noisy alerts that non-expert facility managers struggle to use. This paper proposes an end-to-end agentic pipeline that combines deep time-series forecasting, variational anomaly detection, and LLM-based reasoning to generate prioritized, actionable maintenance recommendations. The system tracks seven office appliances using a hybrid Singular Spectrum Analysis (SSA) and Long Short-Term Memory (LSTM) forecasting model, and applies a per-appliance LSTM Variational Autoencoder (VAE) with attention to flag abnormal daily consumption episodes. A three-stage LangChain pipeline begins with a Context Agent that always retrieves three core RAG sources model reliability, hourly baseline, and expert knowledge and conditionally adds up to three more (forecast context, anomaly history, global baseline) based on event characteristics, capped at eight reasoning steps. A Diagnosis Agent converts the evidence into a structured JSON diagnosis, and a Report Agent renders a human-readable narrative. A reflective memory layer incorporates operator feedback. The dashboard shows real-time 30-minute forecasts, intraday consumption, the previous day anomaly report, and a feedback form. We evaluate the forecasting model, anomaly detector with appliance-specific thresholds, and LLM reasoning on a 16-scenario benchmark including sustained and transient spikes, unexpected shutdowns, and systemic events, comparing five LLM backends under static vs. dynamic retrieval. Dynamic retrieval matches full static retrieval across all backends while cutting average context from six to three–six sources per event. The best backend scores 90.4/100 with a 100\% pass rate at a 70-point threshold, and a fully local 7B-parameter model passes all 16 scenarios.
\end{abstract}

\vspace{5pt}
\begin{center}
    \textit{© 2025 IEEE. This is the author’s version of the work accepted for publication in the IEEE International Conference on Communication, Computing, Networking, and Control in Cyber-Physical Systems (CCNCPS 2026). 
    The final version will be available via IEEE Xplore.}
\end{center}
\vspace{5pt}

\section{Introduction}

Buildings account for a substantial share of global electricity demand, and office environments in particular suffer from waste caused by faulty appliances, forgotten devices, and degraded equipment that continue to draw power without being noticed \cite{b_iea}. Sub-metering technologies now make it routine to record energy consumption at the appliance level, yet most facility managers still rely on aggregated dashboards and manual inspection to react to anomalies \cite{b_himeur_survey}. The problem is no longer one of measurement, but one of \emph{interpretation}: a stream of half-hourly readings from a fridge, a coffee machine, and a printer must be turned into a small number of clear, prioritized actions.
Deep learning (DL) approaches for time-series forecasting and anomaly detection now enable reliable identification of unusual appliance behavior without labeled fault data \cite{b_lstm_vae,b_park_lstm_vae}. Hybrid decomposition models such as SSA-LSTM \cite{b_ssa_lstm} improve forecasting accuracy, while LSTM-VAEs with percentile-based thresholds \cite{b_threshold_calib} convert reconstruction errors into appliance-level alerts. However, global thresholds fail for heterogeneous appliance portfolios \cite{b_threshold_calib}, and most systems stop at alert generation, placing the interpretive burden on facility managers who lack the technical expertise to prioritize raw anomaly scores \cite{b_himeur_survey}. Recent work addresses interpretability by encoding energy time series as wavelet or recurrence-plot images and feeding them to vision-LLMs \cite{b_bechar_vllm_wavelet,b_bechar_gaf_mtf}, demonstrating that visual representations improve diagnostic fidelity; however, these end-to-end architectures sacrifice the modularity needed for safety-critical deployments. At the feature extraction level, attention-based transformer representations have shown strong capacity for capturing temporal dependencies in time-series anomaly detection \cite{b_attention}, motivating their use as an interpretable intermediate layer before alert generation.
On the reasoning side, ReAct-style agents \cite{b_react} combine LLM reasoning with external tool use, while retrieval-augmented generation (RAG) \cite{b_rag} grounds outputs in external knowledge bases implemented with FAISS \cite{b_faiss} and lightweight encoders such as MiniLM \cite{b_minilm}. LangChain \cite{b_langchain} supports structured tool orchestration, while recent surveys highlight increasingly sophisticated behaviors of LLM agents \cite{b_llm_industrial}. These agents have been applied to building energy management tasks such as device control, scheduling, and energy analysis \cite{b_llm_smartbuilding}. Adaptive retrieval strategies \cite{b_self_rag} further show that selecting context conditionally based on query characteristics matches full-retrieval accuracy while reducing overhead. Nevertheless, LLMs risk hallucination and overconfident urgency assignments when deployed without deterministic safety nets \cite{b_hallucination}, and agentic pipelines that fix the set of retrieved sources at every call incur unnecessary context overhead for simple events.
These limitations motivate the present work. We propose Smart Energy Agent, a hybrid agentic pipeline that combines (i) an SSA-LSTM forecasting backbone shared across appliances, (ii) per-appliance LSTM-VAE with multi-head attention anomaly detectors with calibrated thresholds, and (iii) a three-stage LLM agent that retrieves three core RAG sources unconditionally and conditionally selects up to three additional sources based on event characteristics, combined with a feedback memory, before producing structured diagnoses and natural-language recommendations. Hard rules and reliability caps surround the LLM to prevent overconfident or unsafe outputs. The contributions of this work are:
\begin{itemize}
    \item A unified architecture coupling deep forecasting, deep anomaly detection, and an LLM agent through retrieval-augmented reasoning, with explicit safety nets on urgency and confidence.
    \item A two-tier retrieval strategy in which the Context Agent always retrieves three core RAG sources (model reliability, hourly baseline, and expert knowledge) and conditionally selects up to three additional sources (forecast context, anomaly history, and global baseline) based on event characteristics, reducing average context size from 6 to 3--6 sources per event.
    \item A per-appliance threshold calibration strategy that adapts detector strictness to the noise profile of each device.
    \item A 16-scenario evaluation benchmark with a 100-point rubric covering urgency classification, likely cause, confidence calibration, special-case flags, and actionability, used to compare dynamic versus static retrieval across five LLM backends, confirming equivalent diagnostic accuracy with reduced context overhead.
    \item A comparison of five LLM backends including a fully local 7B-parameter model, confirming pipeline viability without external API dependencies.
\end{itemize}
\section{Methodology}
\begin{figure*}[!t]
\centering
\includegraphics[width=0.80\textwidth]{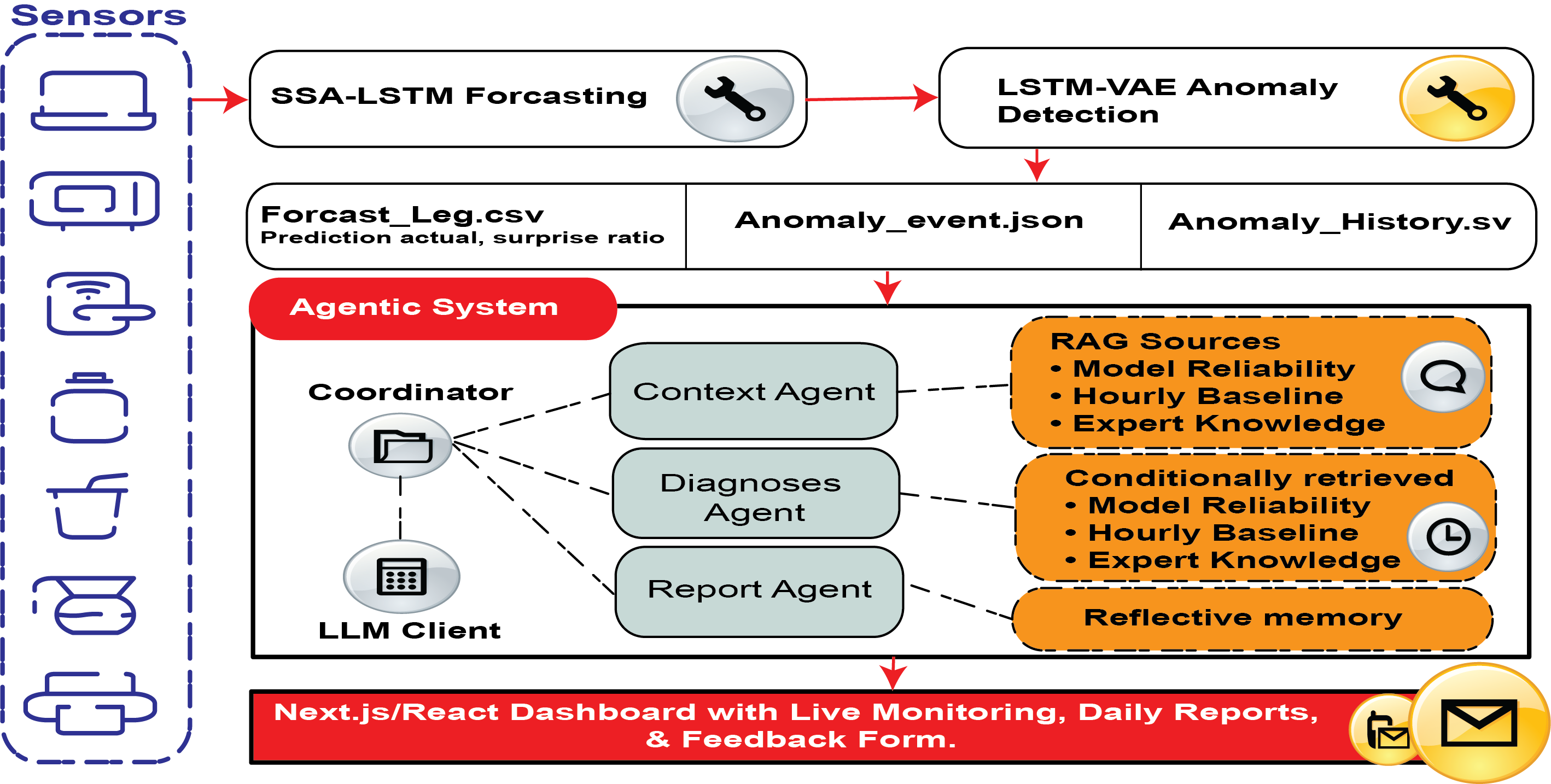}
\caption{End-to-end architecture of the Smart Energy Agent.}
\label{fig:architecture}
\end{figure*}

Fig.~\ref{fig:architecture} summarizes the end-to-end architecture of the proposed system, from raw sensor readings through forecasting, anomaly detection, and agentic reasoning to the operator dashboard. The architecture operates at two 11timescales. Every 30 minutes, sensor readings feed the forecasting module, which appends predictions and actuals to \texttt{forecast\_log.csv} and updates the live monitoring view. Every 24 hours, the anomaly detector scores all appliances, overwrites \texttt{anomaly\_events.json} with enriched event objects, and appends new flags to \texttt{anomaly\_history.csv}; the Context Agent, Diagnosis Agent, and Report Agent then consume these outputs to produce the daily report.
\subsection{Sensors and Data Layer}
Seven appliances are monitored — a coffee machine, a fridge, a microwave, a kettle, a printer, a water dispenser, and a Tasmota-based smart plug — each sampled every 30 minutes as a (timestamp, appliance, kWh) tuple. Four persistent stores are maintained: \texttt{raw\_readings.csv}, \texttt{forecast\_log.csv} (seven rows are appended for each appliance per forecasting run), \texttt{anomaly\_history.csv} (a rolling 30-day flag record updated nightly), \texttt{anomaly\_events.json} (enriched event objects overwritten each nightly run), and \texttt{feedback\_log.csv} (operator corrections).
\subsection{Component Overview}
\begin{itemize}
    \item \textbf{Forecasting Model:} SSA-LSTM hybrid producing the next 30-minute prediction per appliance; outputs a \emph{surprise ratio} used downstream.
    \item \textbf{Anomaly Detector:} Seven independent LSTM-VAE models with multi-head attention and percentile-based thresholds.
    \item \textbf{Coordinator:} Groups events by time and appliance, filters borderline cases, and caps LLM calls per run.
    \item \textbf{Context Agent:} Tool-calling executor with up to eight reasoning iterations; applies two-tier RAG retrieval.
    \item \textbf{Diagnosis Agent:} Applies hybrid rule + LLM logic to produce a structured JSON diagnosis.
    \item \textbf{Report Agent:} Converts JSON diagnoses into a prioritized natural-language report.
    \item \textbf{Feedback Memory:} Reflective store of past operator corrections injected into future diagnosis prompts.
    \item \textbf{Dashboard:} Next.js/React application with live monitoring, daily report, and feedback form views.
\end{itemize}
\subsection{Forecasting with SSA-LSTM}
The forecasting backbone is a single global model with separate LSTM branches for each SSA component (trend, seasonality, residual) and an additional branch for exogenous features. A learned per-appliance embedding allows specialization without per-device training. Inputs include lagged consumption values, rolling statistics, and cyclic time encodings. The model is trained once on a chronological 70/15/15 split and deployed in inference-only mode every 30 minutes.
\subsection{Anomaly Detection with LSTM-VAE}
For each appliance we train a separate LSTM-VAE: a bidirectional LSTM encoder $f_\phi$ with multi-head attention produces an approximate posterior $q_\phi(z\mid x)=\mathcal{N}(\mu_\phi(x),\sigma_\phi^2(x))$, and an LSTM decoder $g_\theta$ reconstructs the input window $x \in \mathbb{R}^{T \times F}$. Training minimizes a feature-weighted ELBO,
\begin{equation}
\begin{split}
\mathcal{L}(\phi,\theta;x)
  &= \mathbb{E}_{q_\phi(z|x)}\!\left[
      \tfrac{1}{TF}\textstyle\sum_{t,f} w_f\bigl(x_{t,f}-g_\theta(z)_{t,f}\bigr)^2
    \right] \\
  &\quad+ \beta\,D_{\mathrm{KL}}\!\left(q_\phi(z|x)\,\|\,\mathcal{N}(0,I)\right),
\end{split}
\label{eq:elbo}
\end{equation}
where $w_f$ up-weights energy features (e.g.\ $w_{\mathrm{kWh}}{=}10$) and $\beta$ is annealed from $0$ to $0.05$ to prevent posterior collapse. At inference, a sliding window scores each timestep by averaging the weighted reconstruction error over all overlapping windows that contain it, using $\mu_\phi$ as a deterministic latent code,
\begin{equation}
s(t) = \frac{1}{|\mathcal{W}(t)|}\sum_{w\in\mathcal{W}(t)}
       \tfrac{1}{F}\textstyle\sum_f w_f\bigl(x_{t,f}-g_\theta(\mu_\phi(x^{(w)}))_{t,f}\bigr)^2.
\label{eq:score}
\end{equation}
The detection threshold $\tau$ is the $p$-th percentile of validation scores (chronological middle 15\%, used solely for calibration),
\begin{equation}
\tau = Q_p\!\left(\{s(t):t\in\mathcal{V}\}\right),
\label{eq:threshold}
\end{equation}
with $p$ tuned per appliance: Microwave 97.0\%, Printer 98.5\%, Fridge/Kettle/Coffee Machine 99.0\%, Water Dispenser 99.5\%, and Tasmota 99.9\%. Noisier devices receive lower thresholds to preserve sensitivity; more predictable ones receive stricter thresholds since a high reconstruction error is genuinely surprising. Consecutive flagged timesteps are merged into events and enriched with energy statistics, severity, and pattern type.
\subsection{Agentic Pipeline}
The pipeline comprises three sequential stages: a Context Agent for evidence retrieval, a Diagnosis Agent for interpretation, and a Report Agent for narrative generation.
The Context Agent is a LangChain tool-calling executor with six registered tools:
\begin{enumerate}
    \item \texttt{get\_expert\_knowledge()} -- semantic search over a FAISS-backed appliance knowledge base using \textit{all-MiniLM-L6-v2} embeddings.
    \item \texttt{get\_model\_reliability()} -- the false-alarm profile of the detector for that appliance.
    \item \texttt{get\_hourly\_baseline()} -- expected consumption for the specific hour and weekday.
    \item \texttt{get\_global\_baseline()} -- training-set energy statistics for the appliance.
    \item \texttt{get\_anomaly\_history()} -- recent anomalies for that appliance.
    \item \texttt{get\_forecast\_context()} -- whether the forecaster was also surprised at the time of the event.
\end{enumerate}
The first three are retrieved unconditionally — the agent must always know detector reliability, the time-of-day baseline, and domain knowledge before reasoning. The remaining three are conditionally selected based on signal ambiguity, event severity, or baseline sparsity, keeping the diagnosis prompt focused and free of irrelevant context.
The Diagnosis Agent routes each event through hybrid rule + LLM logic across three paths: Path~1 returns a deterministic ``inspect now'' diagnosis for devices completely inactive for several days; Path~2 closes the event as normal variation if statistics fall within the historical range; Path~3 invokes the LLM with the full retrieved context, producing a JSON object with eight fields: \texttt{diagnosis}, \texttt{likely\_cause}, \texttt{confidence}, \texttt{confidence\_reason}, \texttt{recommended\_action}, \texttt{urgency}, \texttt{urgency\_reason}, and \texttt{explanation}. Four post-generation safety nets are applied: an urgency fallback, a confidence fallback, a reliability cap that lowers confidence for known noisy detectors, and a night-time hard rule that forces \texttt{immediate} urgency for fridge or water dispenser anomalies between 10~PM and 6~AM. If three or more appliances are simultaneously anomalous, a systemic event rule redirects the diagnosis to a building-level cause.
The Report Agent converts the structured JSON into a prioritized Markdown report grouped by urgency.
\subsection{Feedback Loop and Dashboard}
Operator corrections are appended to \texttt{feedback\_log.csv} via the dashboard feedback form. On each nightly run, corrections matching the current event by appliance, pattern type, and time-of-day window are injected into the diagnosis prompt, providing a reflective memory that progressively reduces repeated false alarms.
The dashboard provides three views: live monitoring (full intraday consumption trace per appliance with the next 30-minute forecast, refreshed each sensor cycle), daily report (yesterday's anomalies rendered as a structured narrative with severity, urgency, likely cause, recommended action, and confidence), and a feedback form for operator corrections.
\section{Experimental Evaluation}
\subsection{Setup}
All experiments are conducted on a chronological train/validation/test split of the appliance-level dataset. The validation set (the middle 15\%) is used solely for threshold calibration; metrics are reported on the held-out test set. The forecasting model uses 30-minute granularity. The LLM stage is evaluated separately on a benchmark of 16 hand-crafted scenarios (S01--S16) covering sustained spikes, transient spikes, unexpected shutdowns, extended inactivity, high-variability episodes, and systemic events across all seven appliances.
\subsection{Forecasting Performance}
Table~\ref{tab:forecast_global} reports the global metrics across all appliances. The model achieves an $R^2$ of 0.9976 and a Weighted Absolute Percentage Error (WAPE) of 1.32\%, indicating that variance is well captured at the aggregate level. Table~\ref{tab:forecast_appliance} breaks results down per appliance.

\begin{table}[!t]
\caption{Global Forecasting Metrics on the Test Set}
\label{tab:forecast_global}
\centering
\renewcommand{\arraystretch}{1.35}
\setlength{\tabcolsep}{4pt}
\begin{tabular*}{\columnwidth}{@{\extracolsep{\fill}}lr@{}}
\toprule
\textbf{Metric} & \textbf{Value} \\
\midrule
MAE  (kWh) & 3.04 \\
RMSE (kWh) & 15.40 \\
WAPE (\%)  & 1.32 \\
$R^{2}$    & 0.9976 \\
\bottomrule
\end{tabular*}
\end{table}

\begin{table}[!t]
\caption{Per-Appliance Forecasting Metrics on the Test Set}
\label{tab:forecast_appliance}
\centering
\renewcommand{\arraystretch}{1.3}
\setlength{\tabcolsep}{4pt}
\begin{tabular*}{\columnwidth}{@{\extracolsep{\fill}}lrrr@{}}
\toprule
\textbf{Appliance} & \textbf{MAE (kWh)} & \textbf{RMSE (kWh)} & \textbf{WAPE (\%)} \\
\midrule
Coffee machine  & 3.34 & 14.80 & 6.2 \\
Fridge          & 0.14 &  0.16 & 0.0 \\
Kettle          & 5.35 & 15.42 & 3.3 \\
Microwave       & 5.67 & 11.78 & 21.2 \\
Printer         & 0.03 &  0.05 & 0.0 \\
Tasmota         & 1.03 &  7.13 & 10.8 \\
Water dispenser & 5.71 & 31.84 & 5.9 \\
\bottomrule
\end{tabular*}
\end{table}

The fridge and the printer, both with stereotypical patterns, are forecast almost perfectly. The microwave shows the highest relative error (WAPE = 21.2\%), consistent with its bursty on/off behavior. The water dispenser exhibits the largest absolute RMSE (31.84 kWh), driven by infrequent but intense heating spikes. These patterns motivate the appliance-specific threshold calibration of the anomaly detector.
\subsection{Anomaly Detection Reliability}
Table~\ref{tab:anomaly_reliability} summarizes the reliability profile of each LSTM-VAE model. Reliability is a qualitative trust assessment derived from threshold strictness, validation behavior, and observed false-alarm rates over 30 days of operation. Devices marked LOWER such as the microwave are flagged frequently and require corroborating evidence; devices marked HIGH (fridge, printer) are trusted by default when they raise an alert.

\begin{table}[!t]
\caption{Per-Appliance Detector Reliability and Threshold}
\label{tab:anomaly_reliability}
\centering
\renewcommand{\arraystretch}{1.3}
\setlength{\tabcolsep}{4pt}
\begin{tabular*}{\columnwidth}{@{\extracolsep{\fill}}lcr@{}}
\toprule
\textbf{Appliance} & \textbf{Reliability} & \textbf{Threshold (\%)} \\
\midrule
Fridge          & HIGH     & 99.0 \\
Printer         & HIGH     & 99.5 \\
Coffee machine  & GOOD     & 99.0 \\
Kettle          & GOOD     & 99.0 \\
Tasmota         & MODERATE & 99.9 \\
Water dispenser & MODERATE & 99.5 \\
Microwave       & LOWER    & 97.0 \\
\bottomrule
\end{tabular*}
\end{table}

\subsection{LLM Backend Comparison}
The LLM reasoning stage is evaluated using a 100-point rubric per scenario, broken down as follows:
\begin{itemize}
    \item Urgency classification (35 pts): full / partial / zero match against the expected level.
    \item Likely cause keywords (25 pts): proportional, synonym-aware coverage of the expected cause.
    \item Confidence level (20 pts): match between stated confidence and underlying model reliability.
    \item Special flags (10 pts): correct flagging of night-time, concurrent, or reliability-related caveats.
    \item Actionability (10 pts): presence of a concrete action verb and target object.
\end{itemize}
A scenario is considered \emph{passed} when its score reaches 70/100 or higher.
We tested five backends from open-source (Ollama) and commercial APIs (Anthropic, OpenRouter). Each was evaluated under both static retrieval (all six RAG sources retrieved unconditionally) and dynamic retrieval (three core sources always retrieved, up to three additional selected conditionally). Table~\ref{tab:llm_results} shows dynamic retrieval results, which matched static scores exactly across all backends and scenarios.

\begin{table}[!t]
\caption{Scenario-Based Evaluation of Five LLM Backends (16 Scenarios, Dynamic Retrieval)}
\label{tab:llm_results}
\centering
\renewcommand{\arraystretch}{1.3}
\setlength{\tabcolsep}{4pt}
\begin{tabular*}{\columnwidth}{@{\extracolsep{\fill}}llrc@{}}
\toprule
\textbf{Model} & \textbf{Backend} & \textbf{Avg Score} & \textbf{Passed} \\
\midrule
deepseek-v4-pro & DeepSeek API    & 90.4 & 16/16 \\
gpt-5.5-pro     & OpenRouter API  & 88.9 & 16/16 \\
gemini-3.1-pro  & Gemini API      & 88.8 & 15/16 \\
claude-sonnet-4 & Anthropic API   & 88.5 & 16/16 \\
qwen2.5:7b      & Ollama (local)  & 85.4 & 16/16 \\
\bottomrule
\end{tabular*}
\end{table}

Three observations stand out. First, hosted models cluster within a 1.9-point range (88.5--90.4), suggesting frontier model choice has limited impact on diagnosis quality given sufficient retrieval context. Second, all hosted models pass at least 15/16 scenarios, with only one borderline case separating Gemini. Third, the local 7B Qwen2.5 passes all 16 scenarios at 85.4---roughly five points below the best hosted model but well above threshold, confirming API-free feasibility.
Dynamic retrieval matched full static retrieval accuracy while reducing average context from 6 to 3--6 sources per event. Although retrieval cost differences are negligible in local RAG setups, dynamic selection becomes cost-justified in production environments with API-based or rate-limited sources, where skipping 2--3 retrievals per event compounds significantly across thousands of daily monitoring cycles.
\subsection{Component Ablation}
The hybrid routing logic plays a measurable role in robustness. Removing the night-time hard rule, the reliability cap, or the rule-based urgency fallback caused intermittent failures during exploratory development, particularly on rare or under-represented event types. The deterministic Path 1 short-circuit (devices completely off for several days) further reduces unnecessary LLM calls and avoids spurious confidence values on what is, in fact, a certainty.

\subsection{Comparison with Related Work on LLM Reasoning Evaluation}
Table~\ref{tab:related_comparison} maps the three most relevant cited systems
onto a common set of dimensions. Two observations follow.
The two closest architectural relatives~\cite{b_bechar_vllm_wavelet,b_bechar_gaf_mtf}
evaluate LLM output via validation loss and AHP scoring respectively — neither
decomposes diagnostic quality along the operationally critical dimensions of
urgency correctness, confidence calibration, or actionability that our rubric
captures. Their results confirm that visual representations improve detection
fidelity, but neither provides a pass/fail diagnostic benchmark comparable
to ours.
Finally, RAGAS~\cite{b_ragas} and ARES~\cite{b_ares} are evaluation
\emph{frameworks} rather than systems, and both confirm that generic
automated metrics are insufficient for domain-specific diagnostic tasks ---
directly motivating the hand-crafted 100-point rubric used here.

\begin{table*}[!t]
\caption{Comparison of LLM Reasoning Evaluation Across Related Work}
\label{tab:related_comparison}
\centering
\renewcommand{\arraystretch}{1.3}
\setlength{\tabcolsep}{5pt}
\begin{tabularx}{\textwidth}{@{}>{\raggedright\arraybackslash}p{2.1cm}
                                  >{\raggedright\arraybackslash}X
                                  >{\raggedright\arraybackslash}p{2.6cm}
                                  >{\raggedright\arraybackslash}X
                                  >{\raggedright\arraybackslash}p{2.3cm}
                                  >{\raggedright\arraybackslash}X@{}}
\toprule
\textbf{Work} &
\textbf{Task scope} &
\textbf{Model} &
\textbf{Evaluation protocol} &
\textbf{Metric} &
\textbf{Key result} \\
\midrule
Ours (Smart Energy Agent) &
Appliance-level fault diagnosis \& prioritised recommendations &
5 LLM backends incl.\ local 7B (no fine-tuning) &
16-scenario hand-crafted benchmark; 100-pt rubric (urgency 35, cause 25, confidence 20, flags 10, actionability 10) &
Avg.\ score /100; pass rate ($\geq$70) &
Best hosted: 90.4/100, 16/16 pass; local 7B: 85.4/100, 16/16 pass \\
\addlinespace
Bechar et al.\ \cite{b_bechar_vllm_wavelet} (ICDM 2025) &
Anomaly detection + optimisation recommendations from building energy data &
Idefics-7B VLLM (fine-tuned on CWT/RP visual encodings) &
Supervised fine-tuning on Univ.\ of Sharjah dataset; held-out validation split &
Validation loss $\downarrow$ &
CWT: 0.0952; RP: 0.1064; raw TS baseline: 0.1176 \\
\addlinespace
Bechar et al.\ \cite{b_bechar_gaf_mtf} (Expert Syst.\ Appl., 2026) &
Energy consumption pattern decoding + report generation (GAF/MTF visual encodings) &
VLLM on 3-D GAF \& MTF charts &
AHP-based multi-evaluator framework; comparison across 1-D, 2-D, 3-D representations &
AHP score; ROUGE-1 &
3-D GAF/MTF: AHP 0.659, ROUGE-1 0.4408; 2-D: 0.642; 1-D: 0.594 \\
\addlinespace
RAGAS \cite{b_ragas} (EACL 2024) &
General-purpose RAG evaluation (open-domain QA) &
Framework (various LLMs as judges) &
Reference-free; four automatic metrics on open-domain QA sets &
Faithfulness, answer relevancy, context precision/recall (0--1) &
Evaluation framework; no single target score — motivates domain-specific rubrics \\
\addlinespace
ARES \cite{b_ares} (NAACL 2024) &
General-purpose RAG evaluation (8 tasks: KILT, SuperGLUE, AIS) &
Fine-tuned lightweight LM judges + PPI; few hundred human annotations &
Synthetic training data; prediction-powered inference; domain-shift tests &
Kendall's $\tau$ ranking accuracy vs.\ RAGAS &
$+$0.065 on context relevance; $+$0.132 on answer relevance vs.\ RAGAS — advocates task-specific rubrics \\
\bottomrule
\end{tabularx}
\smallskip

\noindent\footnotesize
\textit{Notes}: ``Validation loss'' (lower $\downarrow$ is better) and ``AHP score'' (higher $\uparrow$ is better)
are not directly comparable to the rubric score used in this work; they are included to characterise
each system's evaluation methodology.
\end{table*}

\section{Discussion}
The most effective design choice was separating numerical evidence gathering from language generation: forecasting and anomaly detection produce structured artifacts; the Context Agent assembles a focused briefing; the Diagnosis Agent emits strict JSON; the Report Agent renders the narrative. This localizes errors — a misformatted JSON does not corrupt the report, and a poorly worded report does not affect the structured event log.

Per-appliance thresholds yielded a clear payoff over a single global threshold, which either flooded operators with microwave alerts or suppressed genuine fridge anomalies. Dynamic context selection confirmed that diagnostic performance was identical whether three or six sources were retrieved, validating focused retrieval — a benefit expected to compound in production environments with rate-limited external APIs. The reflective feedback memory is fully integrated, injecting operator corrections into future prompts when a matching event is found by appliance, pattern type, and time of day; quantifying its impact on accuracy requires sustained expert-validated corrections and is left for future work. The 16-scenario benchmark covers the operationally relevant event types observed over 30 days; a larger adversarial benchmark would further strengthen evaluation. The full stack runs on free-tier infrastructure: FAISS with \textit{all-MiniLM-L6-v2} on CPU, with an optional local Qwen2.5 backend for on-premises data residency.
\section{Conclusion}
We presented Smart Energy Agent, a hybrid pipeline combining deep forecasting, anomaly detection, retrieval-augmented LLM reasoning, and operator feedback for appliance-level energy management. Separating numerical evidence gathering from language generation and surrounding the LLM with deterministic safety nets yields reliable, prioritized recommendations even when detectors are noisy. Forecasting reaches $R^2 = 0.9976$ on the held-out test set; the agentic stage scores 90.4/100 on a 16-scenario benchmark with the best hosted backend, while a fully local 7B model passes all 16 scenarios — confirming viability without external API dependencies. Future work will focus on three directions: extending the benchmark with adversarial scenarios to stress-test the safety nets, quantifying the impact of the reflective feedback memory on repeated false-alarm reduction over time, and validating dynamic retrieval cost savings under production-scale deployments with external knowledge sources.
\section*{Acknowledgment}
The authors thank the team at Laboratoire LITAN, ESTIN, for ongoing
discussions on agentic AI systems for energy management. The authors also
thank the College of Computing and Informatics (CCI) at the University of
Sharjah and the College of Engineering and Information Technology (CEIT) at
the University of Dubai for their support and collaboration.

\end{document}